\documentclass[10pt,twocolumn,letterpaper]{article}

\usepackage{cvpr}
\usepackage{times}
\usepackage{epsfig}
\usepackage{graphicx}
\usepackage{amsmath}
\usepackage{amssymb}
\usepackage{bm}
\usepackage{lipsum}
\usepackage{balance}

\usepackage{placeins} 
\usepackage{flafter}  

\usepackage{tabularx, booktabs}
\usepackage{multirow}
\newcolumntype{Y}{>{\centering\arraybackslash}X}

\usepackage[linesnumbered,ruled,vlined]{algorithm2e}

\SetCommentSty{mycommfont}
\SetKwInput{KwInput}{Input}                
\SetKwInput{KwOutput}{Output}             

\DeclareMathOperator*{\argmin}{\arg\!\min}

\usepackage[pagebackref=true,breaklinks=true,letterpaper=true,colorlinks,bookmarks=false]{hyperref}

\cvprfinalcopy 

\def\cvprPaperID{****} 

\ifcvprfinal\fi
\begin{document}

\title{Robust Single Rotation Averaging\vspace{-0.5em}}

\author{Seong Hun Lee \hspace{25pt} Javier Civera\thanks{This work was partially supported by the Spanish government (project PGC2018-096367-B-I00) and the Arag{\'{o}}n regional government (Grupo DGA-T45{\_}17R/FSE).} \\
I3A, University of Zaragoza, Spain\\
{\tt\small \{seonghunlee, jcivera\}@unizar.es}
}
\maketitle
\thispagestyle{empty}

\begin{abstract}
\vspace{-0.5em}
We propose a novel method for single rotation averaging using the Weiszfeld algorithm.
Our contribution is threefold:
First, we propose a robust initialization based on the elementwise median of the input rotation matrices.
Our initial solution is more accurate and robust than the commonly used chordal $L_2$-mean. 
Second, we propose an outlier rejection scheme that can be incorporated in the Weiszfeld algorithm to improve the robustness of $L_1$ rotation averaging.
Third, we propose a method for approximating the chordal $L_1$-mean using the Weiszfeld algorithm.
An extensive evaluation shows that both our method and the state of the art perform equally well with the proposed outlier rejection scheme, but ours is $2-4$ times faster.
\end{abstract}

\vspace{-1em}
\section{Introduction}
We consider the problem of single rotation averaging, i.e., averaging several estimates of a single rotation to obtain the best estimate.
This problem is relevant in many applications such as structure from motion (SfM) \cite{hartley2011L1, tron2016survey}, camera rig calibration \cite{dai2009rotation}, motion capture \cite{inna2010arithmetic}, satellite/spacecraft attitude determination \cite{lam2007precision,markley2007averaging} and crystallography \cite{humbert1996determination, morawiec1998note}. 

A standard approach for single rotation averaging is to find the rotation that minimizes a cost function based on the distance to the input rotations.
We refer to \cite{hartley2013rotation} for an extensive study of various distance functions.
The current state-of-the-art method is to minimize the sum of geodesic distances using the Weiszfeld algorithm on $SO(3)$ \cite{hartley2011L1}.

In this work, we propose a novel method, also based on the Weiszfeld algorithm \cite{weiszfeld1,weiszfeld2}, that is faster and more robust than \cite{hartley2011L1}.
Our contributions are as follows:
\begin{enumerate}\itemsep0em
    \item A robust initialization from the elementwise median of the input rotation matrices (Section \ref{subsec:init}).
    \item An implicit outlier rejection scheme performed at each iteration of the Weiszfeld algorithm (Section \ref{subsec:outlier}).
    \item An approximation of the chordal median in $SO(3)$ using the Weiszfeld algorithm (Section \ref{subsec:chordal}).
\end{enumerate}
We substantiate our claim through extensive evaluation on synthetic data (Section \ref{sec:result}).
To download our Matlab code, go to \url{http://seonghun-lee.github.io}.

\section{Preliminaries}
We denote the vectorization of an $n\times m$ matrix by $\mathrm{vec}(\cdot)$ and its inverse by $\mathrm{vec}^{-1}_{n\times m}(\cdot)$.
For a 3D vector $\mathbf{v}$, we define $\mathbf{v}^\wedge$ as the corresponding $3\times3$ skew-symmetric matrix, and denote the inverse operator by $(\cdot)^\vee$, i.e., $\left(\mathbf{v}^\wedge\right)^\vee=\mathbf{v}$.
The Euclidean, the $L_1$ and the Frobenius norm are respectively denoted by $\lVert \cdot \rVert$, $\lVert\cdot\rVert_1$ and $\lVert \cdot \rVert_F$.
A rotation can be represented by a rotation matrix $\mathbf{R}\in SO(3)$ or a rotation vector $\mathbf{v}=\theta\hat{\mathbf{v}}$ where $\theta$ and $\hat{\mathbf{v}}$ are the angle and the unit axis of the rotation, respectively.
The two representations are related by Rodrigues formula, and we denote the corresponding mapping between them by $\text{Exp}(\cdot)$ and $\text{Log}(\cdot)$ \cite{forster2017onmanifold}:
\begin{equation}
    \mathbf{R}=\mathrm{Exp}(\mathbf{v})
    :=
    \mathbf{I}+\frac{\sin\left(\lVert\mathbf{v}\rVert\right)}{\lVert\mathbf{v}\rVert}\mathbf{v}^\wedge+\frac{1-\cos\left(\lVert\mathbf{v}\rVert\right)}{\lVert\mathbf{v}\rVert^2}\left(\mathbf{v}^\wedge\right)^2,
\end{equation}
\begin{equation}
\label{eq:log_map1}
    \mathbf{v}=\mathrm{Log}(\mathbf{R}):=
    \frac{\theta}{2\sin(\theta)}\left(\mathbf{R}-\mathbf{R}^\top\right)^\vee 
\end{equation}
\begin{equation}
\label{eq:log_map2}
    \text{with} \quad \theta = \cos^{-1}\left(\frac{\mathrm{tr}(\mathbf{R})-1  }{2}\right).
\end{equation}
The geodesic distance between two rotations $d_\angle(\mathbf{R}_1, \mathbf{R}_2)$ is obtained by substituting $\mathbf{R}_1\mathbf{R}_2^\top$ into $\mathbf{R}$ in \eqref{eq:log_map2}.
In \cite{hartley2013rotation}, it was shown that the chordal distance is related to the geodesic distance by the following equation:
\begin{align}
    d_\text{chord}(\mathbf{R}_1, \mathbf{R}_2):=& \lVert\mathbf{R}_1-\mathbf{R}_2\rVert_F\\
    =&2\sqrt{2}\sin\left(d_\angle(\mathbf{R}_1, \mathbf{R}_2)/2\right) \label{eq:chordal}.
\end{align}
We define $\mathrm{proj}_{SO(3)}(\cdot)$ as the projection of the $3\times3$ matrix onto the special orthogonal group $SO(3)$, which gives the closest rotation in the Frobenius norm \cite{arun}:
For $\mathbf{M}\in\mathbb{R}^{3\times3}$,
\begin{equation}
\label{eq:orthogonal_projection}
    \mathrm{proj}_{SO(3)}(\mathbf{M}):=\mathbf{UWV}^\top,
\end{equation}
where
\begin{align}
    \mathbf{U}\bm{\Sigma}\mathbf{V}^\top 
    &= \text{SVD}\left(\mathbf{M}\right),  \\
    \mathbf{W}
    &= 
    \begin{cases}
    \text{diag}(1,1,-1)  & \text{if} \ \  \det\left(\mathbf{UV}^\top\right) < 0\\
    \mathbf{I}_{3\times3} & \text{otherwise}
    \end{cases}.
\end{align}

\newpage
\section{Method}
\subsection{Robust Initialization}
\label{subsec:init}
\begin{figure}[t]
 \centering
 \includegraphics[width=0.33\textwidth]{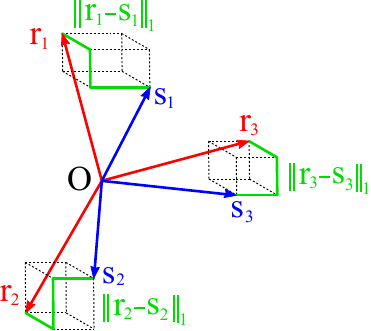}
\caption{The elementwise $L_1$ norm of $(\mathbf{R}-\mathbf{S})$ is equal to $\sum_{i=1}^3\lVert\mathbf{r}_i-\mathbf{s}_i\rVert_1$ where $\mathbf{R}=[\mathbf{r}_1,\mathbf{r}_2,\mathbf{r}_3]$ and $\mathbf{S}=[\mathbf{s}_1,\mathbf{s}_2,\mathbf{s}_3]$.
This can be thought as the total length of the green lines.
 }
\label{fig:entryL1}
\end{figure}

In \cite{hartley2011L1}, the chordal $L_2$ mean of the rotations is taken as the starting point of the Weiszfeld algorithm.
For input rotations $\{\mathbf{R}_i\}_{i=1}^N$, it is given by $\text{proj}_{SO(3)}\left(\sum_{i=1}^N\mathbf{R}_i\right)$ \cite{hartley2013rotation}.
Although this initial solution can be obtained very fast, it is often inaccurate and sensitive to outliers. 
To overcome this weakness, we propose to initialize using the following matrix:
\begin{equation}
    \mathbf{S}_0=\argmin_{\textstyle\mathbf{S}\in\mathbb{R}^{3\times3}} \sum_{i=1}^N\sum_{j=1}^3\sum_{k=1}^3  \left|\left(\mathbf{R}_i-\mathbf{S}\right)_{jk}\right|,
\end{equation}
where the subscript $jk$ denote the element at the $j$-th row and the $k$-th column of the matrix.   
Note that $\sum_{j,k}|\mathbf{M}_{jk}|$ is called the elementwise $L_1$ norm of the matrix $\mathbf{M}$.
See Fig. \ref{fig:entryL1} for the geometric interpretation of this distance metric.
Since the nine entries of $\mathbf{S}$ are independent, we can consider them separately in 1D space.
Then, the entry of $\mathbf{S}_0$ at location $(j, k)$ minimizes the sum of absolute deviations from the entries of $\mathbf{R}_i$'s at $(j, k)$, meaning that it is simply their median:
\begin{equation}
    \left(\mathbf{S}_0\right)_{jk} = \text{median}\left(\{\left(\mathbf{R}_i\right)_{jk}\}_{i=1}^N\right) \ \ \text{for all}\ \ j,k \in \{1,2,3\}.
\end{equation}
The initial rotation matrix is then be obtained by projecting $\mathbf{S}_0$ onto $SO(3)$:
\begin{equation}
\label{eq:init_proj}
    \mathbf{R}_0 = \mathrm{proj}_{SO(3)}
    \left(\mathbf{S}_0\right).
\end{equation}

\subsection{Outlier Rejection in the Weiszfeld Algorithm}
\label{subsec:outlier}
The geodesic $L_1$-mean (i.e., median) of the rotations is defined as
\begin{equation}
    \mathbf{R}_\mathrm{gm}=\argmin_{\mathbf{R}\in SO(3)} \sum_{i=1}^N d_\angle(\mathbf{R}_i, \mathbf{R}).
\end{equation}
In \cite{hartley2011L1}, it was shown that this can be computed using the Weiszfeld algorithm on $SO(3)$ and that it is more robust to outliers than the $L_2$-mean.
However, a large number of outliers is still critical to the accuracy.
To further mitigate the influence of the outliers, we modify the Weiszfeld algorithm of \cite{hartley2011L1} such that the large residuals are given zero weight at each iteration.
Specifically, we disregard all the residuals larger than $\text{max}(d_{Q1}, d_\text{max})$ where $d_{Q1}$ is the first quartile of the residuals at each iteration and $d_\text{max}$ is some threshold we set in order to avoid discarding inliers.
The details are given in Algorithm \ref{al:geodesic}.
A similar approach of disregarding large residuals was used in \cite{ferraz2014very} for robust Perspective-$n$-Point (PnP) problem.
Note that our approach contrasts \cite{aftab2015convergence} where a smooth robust cost function is favored for theoretically guaranteed convergence.
In practice, our method is more robust to outliers (see Section \ref{subsec:result_comparison}).

\subsection{Approximate Chordal $L_1$-Mean}
\label{subsec:chordal}
The chordal $L_1$-mean of the rotations is defined as 
\begin{equation}
    \mathbf{R}_\mathrm{cm}=\argmin_{\mathbf{R}\in SO(3)} \sum_{i=1}^N d_\text{chord}(\mathbf{R}_i, \mathbf{R}).
\end{equation}
In \cite{hartley2013rotation}, a locally convergent algorithm on $SO(3)$ is proposed for this problem.
In this work, we propose a different approach:
Instead of iteratively updating the estimate on $SO(3)$, we first embed the rotations in a Euclidean space $\mathbb{R}^9$, find their geometric median in $\mathbb{R}^9$ using the standard Weiszfeld algorithm \cite{weiszfeld1,weiszfeld2} (which is globally convergent), and then project this median onto $SO(3)$.
In other words, we approximate $\mathbf{R}_\mathrm{cm}$ as
\begin{equation}
    \mathbf{R}_\mathrm{cm}\approx\mathrm{proj}_{SO(3)}
    \left(\mathbf{S}_\mathrm{cm}\right)
\end{equation}
\begin{align}
\text{with} \ \  \mathbf{S}_\mathrm{cm}&=\argmin_{\textstyle\mathbf{S}\in\mathbb{R}^{3\times3}}\sum_{i=1}^N\lVert\mathbf{R}_i-\mathbf{S}\rVert_F\\
&=\text{vec}^{-1}_{3\times3}\left(\argmin_{\textstyle\mathbf{s}\in\mathbb{R}^9}\sum_{i=1}^N\lVert\text{vec}\left(\mathbf{R}_i\right)-\mathbf{s}\rVert\right).
\end{align}
Since the optimization is performed using the Weiszfeld algorithm, we can also incorporate the initialization and outlier rejection scheme in the previous sections.
Algorithm \ref{al:chordal} summarizes our method.

We point out two things in the implementation:
First, since we do not optimize on $SO(3)$, the initial estimate does not have to come from a rotation, and we omit \eqref{eq:init_proj}.
Second, the threshold $d_\text{max}$ must be scaled appropriately when comparing Algorithm \ref{al:geodesic} and \ref{al:chordal}.
Assuming that $\mathbf{s}\in\mathbb{R}^9$ at each iteration does not vastly differ from an embedding of a rotation in $\mathbb{R}^9$, we convert $d_\text{max}$ from geodesic to chordal using \eqref{eq:chordal}, and vice versa.
This is done in line \ref{al:chordal:thr} of Algorithm \ref{al:chordal}.

\begin{algorithm}[t]
\label{al:geodesic}
\caption{Geodesic median in $SO(3)$ \cite{hartley2011L1} with outlier rejection}
\DontPrintSemicolon
  \KwInput{List of rotation matrices $\{\mathbf{R}_i\}_{i=1}^N$}
  \KwOutput{$\mathbf{R}_\mathrm{gm}$}
  \tcc{Initialize (Section \ref{subsec:init}).}
  $\mathbf{S}_0 \gets \mathbf{0}_{3\times3}$;
  
  $\left(\mathbf{S}_0\right)_{jk} \gets \text{median}\left(\{\left(\mathbf{R}_i\right)_{jk}\}_{i=1}^N\right) \ \  \forall j,k = 1,2,3$;
  
  $\mathbf{R}_0 \gets \mathrm{proj}_{SO(3)}\left(\mathbf{S}_0\right)$; \label{al:geodesic:init}
  
  \tcc{Run the Weiszfeld algorithm on SO(3) \cite{hartley2011L1} with outlier rejection (Section \ref{subsec:outlier}).\hspace{-3em}}
  
  $\mathbf{R}_\mathrm{gm} \gets \mathbf{R}_0$;
  
  \For{$\mathrm{it}=1,2, \cdots,10$}
  {
  \While{$\mathbf{R}_\mathrm{gm}\in\{{R}_i\}_{i=1}^N$} 
    {
     $\mathbf{R}_\mathrm{gm}\gets\mathbf{R}_\text{perturb}\mathbf{R}_\mathrm{gm}$;\hspace{-1em}   \tcp*{Perturb slightly}
    }
  
    $\mathbf{v}_i \gets \mathrm{Log}\left(\mathbf{R}_i\mathbf{R}_\mathrm{gm}^\top\right) \ \forall i=1,\cdots,N$; 
    
    $d_i \gets \lVert \mathbf{v}_i \rVert \ \ \forall i=1,\cdots,N$; 
  
    $d_\text{Q1} \gets Q_1\left(\{d_1, \cdots, d_N\}\right)$; \hspace{-1em}\tcp*{First quartile}
    
    $d_\text{max} \gets \begin{cases}1 & \text{if} \ N \leq 50 \\ 0.5 & \text{otherwise}\end{cases}$
    
    $d_\text{thr} \gets \max\left(d_\text{Q1}, d_\text{max}\right)$;

    $w_i \gets \begin{cases}1 & \text{if} \ d_i \leq d_\text{thr} \\ 0 & \text{otherwise}\end{cases} \ \ \forall i=1,\cdots,N$;

    $\displaystyle \Delta\mathbf{v} \gets \frac{\sum_{i=1}^Nw_i\mathbf{v}_i/d_i}{\sum_{i=1}^Nw_i/d_i}$;
    
    $\mathbf{R}_\mathrm{gm}\gets\mathrm{Exp}(\Delta\mathbf{v})\mathbf{R}_\mathrm{gm}$;
    
    \If{$\lVert\Delta\mathbf{v}\rVert<0.001$}
    {
        break;
    }
  }
  \Return{$\mathbf{R}_\mathrm{gm}$}
\end{algorithm}

\begin{algorithm}[t]
\label{al:chordal}
\caption{Approximate chordal median in $SO(3)$ with outlier rejection}
\DontPrintSemicolon
  \KwInput{List of rotation matrices $\{\mathbf{R}_i\}_{i=1}^N$}
  \KwOutput{$\mathbf{R}_\mathrm{cm}$}
  \tcc{Initialize (Section \ref{subsec:init}).}
  $\mathbf{S}_0 \gets \mathbf{0}_{3\times3}$;
  
  $\left(\mathbf{S}_0\right)_{jk} \gets \text{median}\left(\{\left(\mathbf{R}_i\right)_{jk}\}_{i=1}^N\right) \ \  \forall j,k =1,2,3$;
  
   \tcc{Run the Weiszfeld algorithm in 9D space with outlier rejection (Section \ref{subsec:outlier}).\hspace{-3em}}
  
  $\mathbf{s}_\mathrm{cm} \gets \mathrm{vec}(\mathbf{S}_0)$;
  
  \For{$\mathrm{it}=1,2, \cdots, 10$}
  {
      \While{$\mathbf{s}_\mathrm{cm}\in\{\mathrm{vec}(\mathbf{R}_i)\}_{i=1}^N$} 
    {
     $\mathbf{s}_\mathrm{cm}\gets\mathbf{s}_\mathrm{cm}+\mathcal{U}(0, 0.001)$; 
     \tcp*{Perturb}
    }

    $\mathbf{v}_i \gets \text{vec}\left(\mathbf{R}_i\right)-\mathbf{s}_\mathrm{cm} \ \forall i=1,\cdots,N$; 
    
    $d_i \gets \lVert \mathbf{v}_i \rVert \ \ \forall i=1,\cdots,N$;
  
    $d_\text{Q1} \gets Q_1\left(\{d_1, \cdots, d_N\}\right)$; \hspace{-1em}\tcp*{First quartile}
    
    $d_\text{max} \gets \begin{cases}2\sqrt{2}\sin(1/2)\approx1.356 & \text{if} \ N \leq 50 \\ 2\sqrt{2}\sin(0.5/2)\approx 0.700 & \text{otherwise}\end{cases}$ \label{al:chordal:thr}
    
    $d_\text{thr} \gets \max\left(d_\text{Q1}, d_\text{max}\right)$;

    $w_i \gets \begin{cases}1 & \text{if} \ d_{i} \leq d_\text{thr} \\ 0 & \text{otherwise}\end{cases} \ \ \forall i=1,\cdots,N$;

    $\mathbf{s}_\mathrm{cm, prev}\gets \mathbf{s}_\mathrm{cm}$;

    $\displaystyle \mathbf{s}_\mathrm{cm} \gets \frac{\sum_{i=1}^N w_i\mathbf{v}_i/d_i}{\sum_{i=1}^N w_i/d_i}$;
    
    \If{$\lVert\mathbf{s}_\mathrm{cm}-\mathbf{s}_\mathrm{cm, prev}\rVert<0.001$}
    {
        break;
    }
  }
  
  $\mathbf{R}_\mathrm{cm}=\text{proj}_{SO(3)}(\mathrm{vec}^{-1}_{3\times3}\left(\mathbf{s}_\mathrm{cm}\right))$;
  
  \Return{$\mathbf{R}_\mathrm{cm}$}
\end{algorithm}

\section{Results}
\label{sec:result}
\subsection{Initialization}
\label{subsec:result_init}
For evaluation, we generated a synthetic dataset where the inlier rotations follow a Gaussian distribution with $\sigma=5^\circ$, and the outliers have uniformly distributed angles $\in[0,\pi]$ at random directions.
Fig. \ref{fig:init_accuracy} compares the average accuracy of the proposed initial solution (Section \ref{subsec:init}) and the chordal $L_2$-mean \cite{hartley2013rotation} over 1000 runs.
It can be seen that our solution is significantly better than the chordal $L_2$ mean unless the outlier ratio is extremely high (i.e., above $90\%$).
On average, the $L_2$ chordal method takes 0.37 $\mu$s and ours 0.83 $\mu$s per rotation.
This time difference is insignificant compared to the optimization that follows (see Tab. \ref{tab:timings}).

\subsection{Comparison against \cite{hartley2011L1}}
\label{subsec:result_comparison}
\begin{figure*}[t]
 \centering
 \includegraphics[width=\textwidth]{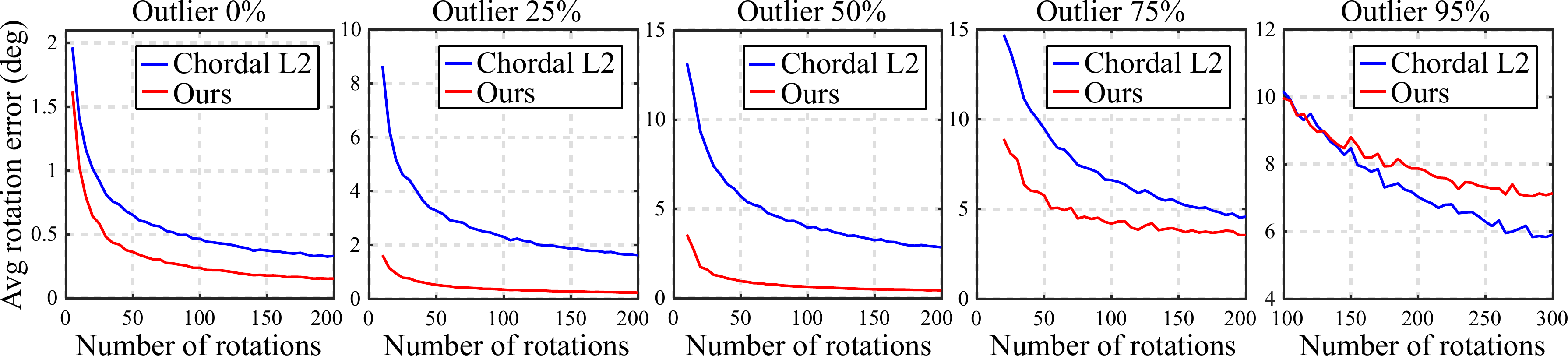}
\caption{Average rotation errors of different initialization methods: Chordal $L_2$-mean \cite{hartley2011L1} versus ours (Section \ref{subsec:init}). 
 }
\label{fig:init_accuracy}
\end{figure*}

\begin{figure*}[t]
 \centering
 \includegraphics[width=\textwidth]{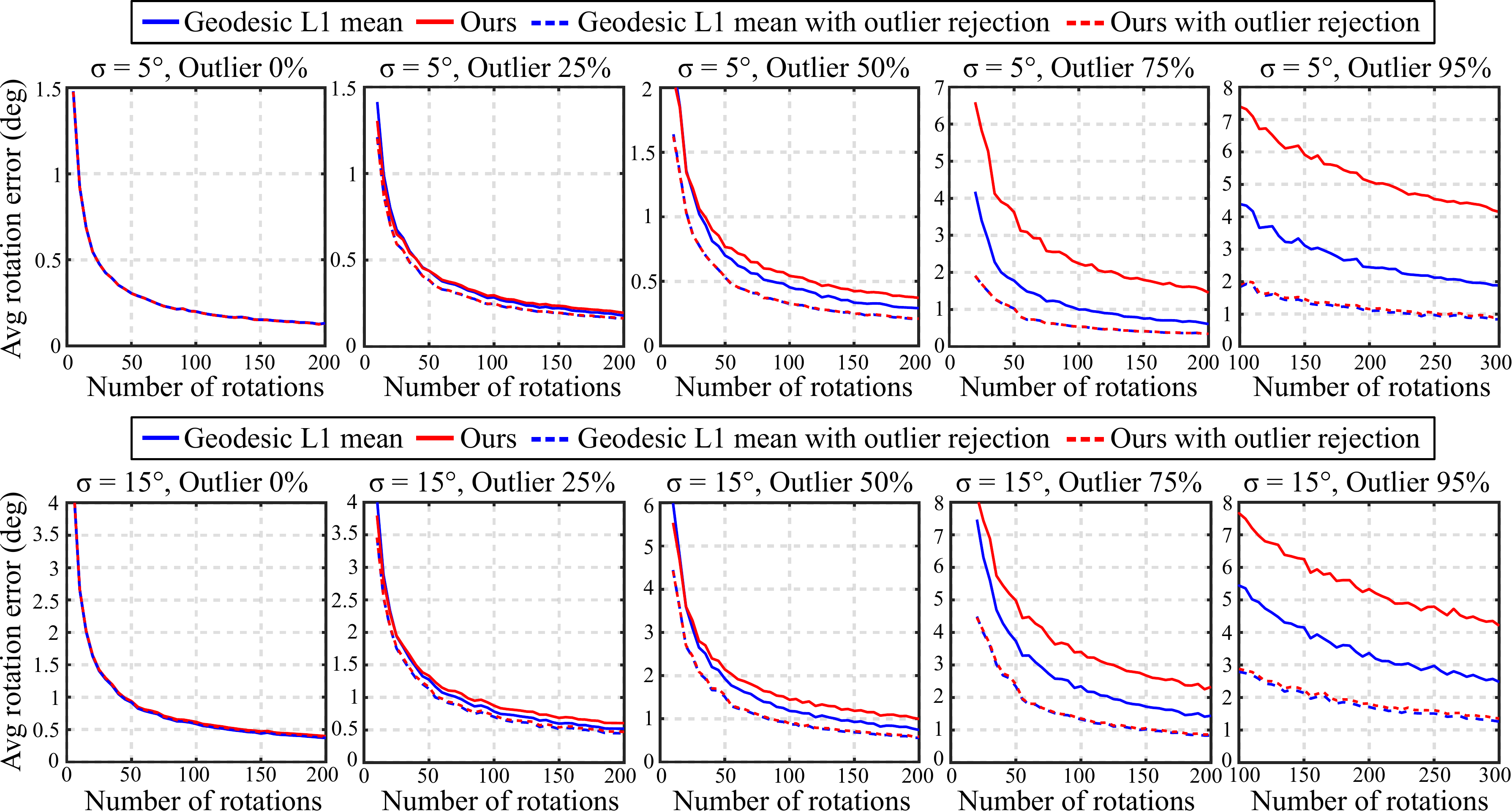}
\caption{Average rotation errors of geodesic $L_1$-mean \cite{hartley2011L1} versus ours (Section \ref{subsec:chordal}), with and without the outlier rejection (Section \ref{subsec:outlier}). 
 }
\label{fig:final_avg}
\end{figure*}

Using the same setup as in previous section, we compare Algorithm \ref{al:geodesic} and \ref{al:chordal}, with and without the proposed outlier rejection scheme (Section \ref{subsec:outlier}).
This time, we consider two different inlier noise levels, $\sigma=5^\circ$ and $15^\circ$. 
The average accuracy of the evaluated methods\footnote{We did not include the chordal $L_2$-mean here, since it produced much larger errors than the rest and was already reported in Fig. \ref{fig:init_accuracy}.} is compared in Fig. \ref{fig:final_avg}.
With the outlier rejection, the geodesic $L_1$-mean and our approximate chordal $L_1$-mean are almost equally accurate. 
Without the outlier rejection, the geodesic $L_1$-mean is more accurate than our approximate chordal $L_1$-mean, but only for very high outlier ratios (i.e, $>50\%$).
Otherwise, there is no significant difference between the two. 

The computation times are reported in Tab. \ref{tab:timings}.
Our method is always faster than \cite{hartley2011L1}, and is 2--4 times faster with the outlier rejection.
That said, the speed is not a major advantage, since all methods can process several hundreds of rotations in less than a millisecond. 
In most cases, averaging rotations will take much less time than other operations, such as the computation of input rotations.

\section{Conclusions}
In this work, we proposed a novel alternative to the work of Hartley et al. \cite{hartley2011L1} for robust single rotation averaging.
While both our method and \cite{hartley2011L1} use the Weiszfeld algorithm, there are three key differences:
\vspace{-0.3em}
\begin{enumerate}\itemsep-0.1em
    \item We initialize the Weiszfeld algorithm using the elementwise median of the input rotation matrices.
    \item We implicitly disregard the outliers at each iteration of the Weiszfeld algorithm.
    \item We approximate the chordal median on $SO(3)$ instead of the geodesic median as in \cite{hartley2011L1}. 
\end{enumerate}
\vspace{-0.3em}
As a result, our method achieves better performance in terms of speed and robustness to outliers.
We also found that incorporating the proposed outlier rejection in the original implementation of \cite{hartley2011L1} leads to similar performance, but at 2--4 times slower speed than ours.

{\renewcommand{\arraystretch}{1.2}%
\vspace{-1em}
\begin{table}[ht]
\small
\begin{center}
\begin{tabular}{c|cc|cc|}
\cline{2-5}
 & \multicolumn{2}{c|}{w/o outlier rejection} & \multicolumn{2}{c|}{w/ outlier rejection} \\
 & \cite{hartley2011L1}& Ours & \cite{hartley2011L1}  & Ours  \\
\hline
\multicolumn{1}{|l|}{($5^\circ$, $0\%$)}& 8.69&  \textbf{4.38} (2.0$\times$) &9.37& \textbf{4.43} (2.1$\times$) \\
\multicolumn{1}{|l|}{($5^\circ$, $25\%$)} & 10.5 & \textbf{4.47} (2.3$\times$) &11.7 & \textbf{5.68} (2.1$\times$)\\
\multicolumn{1}{|l|}{($5^\circ$, $50\%$)} & 15.2 & \textbf{6.21} (2.4$\times$)& 17.2 & \textbf{6.78} (2.5$\times$) \\
\multicolumn{1}{|l|}{($5^\circ$, $75\%$)}& 24.9 & \textbf{15.2} (1.6$\times$) & 27.2 & \textbf{7.71} (3.5$\times$) \\
\multicolumn{1}{|l|}{($5^\circ$, $95\%$)} & 32.1& \textbf{10.3} (3.1$\times$) & 31.6& \textbf{8.99} (3.5$\times$) \\
\multicolumn{1}{|l|}{($15^\circ$, $0\%$)} & 10.7 & \textbf{6.00} (1.8$\times$)& 17.0 & \textbf{6.02} (2.8$\times$) \\
\multicolumn{1}{|l|}{($15^\circ$, $25\%$)} & 15.0 & \textbf{5.98} (2.5$\times$)& 17.0  & \textbf{7.1} (2.4$\times$)  \\
\multicolumn{1}{|l|}{($15^\circ$, $50\%$)} & 19.6 & \textbf{7.66} (2.6$\times$)& 22.3  & \textbf{8.06} (2.8$\times$)  \\
\multicolumn{1}{|l|}{($15^\circ$, $75\%$)} & 24.9& \textbf{11.1} (2.2$\times$)&28.1 & \textbf{8.77} (3.2$\times$)  \\
\multicolumn{1}{|l|}{($15^\circ$, $95\%$)} & 29.1 & \textbf{10.2} (2.9$\times$)&31.7  & \textbf{8.50} (3.7$\times$)  \\
\hline
\end{tabular}
\end{center}
\caption{Median computation time ($\mu s$/rotation) under different inlier noise levels and outlier ratios. The speedup compared to \cite{hartley2011L1} is given in parentheses.
All algorithms were implemented in MATLAB and run on a laptop CPU (Intel i7-4810MQ, 2.8 GHz).}
\label{tab:timings}
\vspace{-5em}
\end{table}
}

\cleardoublepage
{\small
\balance
\bibliographystyle{ieee_fullname}

}

\end{document}